\documentclass{article}
\usepackage{spconf}
\usepackage[utf8]{inputenc} 
\usepackage[T1]{fontenc}    
\usepackage{hyperref}       
\usepackage{url}            
\usepackage{booktabs}       
\usepackage{amsfonts}       
\usepackage{nicefrac}       
\usepackage{microtype}      
\usepackage[algoruled,boxed,lined]{algorithm2e}
\usepackage{graphicx}
\usepackage{pifont}
\usepackage[tight,footnotesize]{subfigure}
\usepackage{amsmath} 
\usepackage{amssymb}  
\usepackage{xcolor}
\usepackage{colortbl}
\usepackage{array,colortbl,ctable}
\usepackage{subfigure}
\newcolumntype{P}[1]{>{\centering\arraybackslash}m{#1}}

\title{UNSUPERVISED DIMENSION SELECTION USING A BLUE NOISE GRAPH SPECTRUM}

\name{Jayaraman J. Thiagarajan$^1$\thanks{This work was performed under the auspices of the U.S. Department of Energy by Lawrence Livermore National Laboratory under Contract DE-AC52-07NA27344.}, Rushil Anirudh$^1$, Rahul Sridhar$^2$ and Peer-Timo Bremer$^1$}
\address{$^1$Lawrence Livermore National Laboratory, $^2$Walmart Labs \\
Email:\{jjayaram@llnl.gov, anirudh1@llnl.gov, rsridha2@uci.edu, bremer5@llnl.gov\}}

\begin{document}
%
\maketitle
\begin{abstract}
Unsupervised dimension selection is an important problem that seeks to reduce dimensionality of data, while preserving the most useful characteristics. While dimensionality reduction is commonly utilized to construct low-dimensional embeddings, they produce feature spaces that are hard to interpret. Further, in applications such as sensor design, one needs to perform reduction directly in the input domain, instead of constructing transformed spaces. Consequently, dimension selection (DS) aims to solve the combinatorial problem of identifying the top-$k$ dimensions, which is required for effective experiment design, reducing data while keeping it interpretable, and designing better sensing mechanisms. In this paper, we develop a novel approach for DS based on graph signal analysis to measure feature influence. By analyzing synthetic graph signals with a blue noise spectrum, we show that we can measure the importance of each dimension. Using experiments in supervised learning and image masking, we demonstrate the superiority of the proposed approach over existing techniques in capturing crucial characteristics of high dimensional spaces, using only a small subset of the original features.
		
\end{abstract}

\section{Introduction}
In this paper, we consider Dimension Selection (DS), which is the problem of selecting the most relevant or influential dimensions from high-dimensional (HD) datasets, such that both the complexity and the robustness of downstream analysis can be improved~\cite{liu2012feature,miller2002subset, zhao2007spectral}. This is crucial in several small data scenarios, where model design becomes more challenging as the dimensionality grows. Though augmenting datasets with more dimensions can be beneficial, undersampling such HD parameter spaces can produce models which rely on noisy correlations. Furthermore, in sensing systems, we often prefer low-dimensional approximations of the high-dimensional data to meet communication, computation, and storage constraints, while retaining the most relevant information~\cite{power}. 

Though feature selection is a well-studied problem in supervisory learning, e.g. lasso~\cite{thiagarajan2014image}, extensions to the unsupervised case have gained a lot of interest. Popular examples include random sampling, principal coordinate analysis~\cite{dadkhahi2015image}, spectral feature identification~\cite{he2006laplacian}, similarity-preserving feature selection~\cite{zhao2013similarity}, and the more recent dimension masking techniques~\cite{dadkhahi2016masking}. Broadly, these approaches rank the different dimensions by their ability to preserve the inherent structure, measured using a variety of spatial and spectral heuristics. In this paper, we propose a novel approach based on graph signal analysis for dimension selection, which is both effective and scalable for high-dimensional data. First, we construct a graph for the dataset, where each node corresponds to a sample, and define a synthetic graph signal characterized by a blue-noise spectrum. Subsequently, we measure the amount of change in the low-frequency content of the signal's spectrum, by perturbing each dimension, based on which we define an importance score. We hypothesize that a feature (dimension) is important if the noisy signal becomes more predictable, i.e. with more energy in the lower frequencies, when that dimension is perturbed. As we show in our  experiments on supervised learning and image masking, this importance score outperforms other existing strategies in selecting relevant dimensions. 



\section{Background - Graph Signal Analysis}
Formally, an undirected weighted graph is represented by the triplet $\mathcal{G} = (\mathcal{V}, \mathcal{E}, \mathbf{A})$, where $\mathcal{V}$ denotes the set of nodes with cardinality $|\mathcal{V}| = N$, $\mathcal{E}$ denotes the set of edges, and $\mathbf{A} \in \mathbb{R}^{N \times N}$ is an adjacency matrix that specifies the weights on the edges, where $\textbf{A}_{i,j}$ corresponds to the edge weight between nodes $v_i$ and $v_j$. Let $\mathcal{N}_{i} = \{j | \mathbf{A}_{i,j} \neq 0\} $ define the neighborhood of node $v_i$, i.e. the set of nodes $v_j$ that have incident edges to it. The normalized graph Laplacian, $\mathbf{L}$, is then constructed as $\mathbf{L} = \mathbf{I} - \mathbf{D}^{-1/2}\mathbf{A} \mathbf{D}^{-1/2},$ where $\mathbf{D}$ is the degree matrix with diagonal entries $\mathbf{D}_{ii} = \sum_{j \in \mathcal{N}_{i}} \mathbf{A}_{i,j}$, and $\mathbf{I}$ denotes the identity matrix.

\begin{figure*}[t]
	\centering
	\subfigure[Spectral domain definition]{\includegraphics[width=0.24\linewidth]{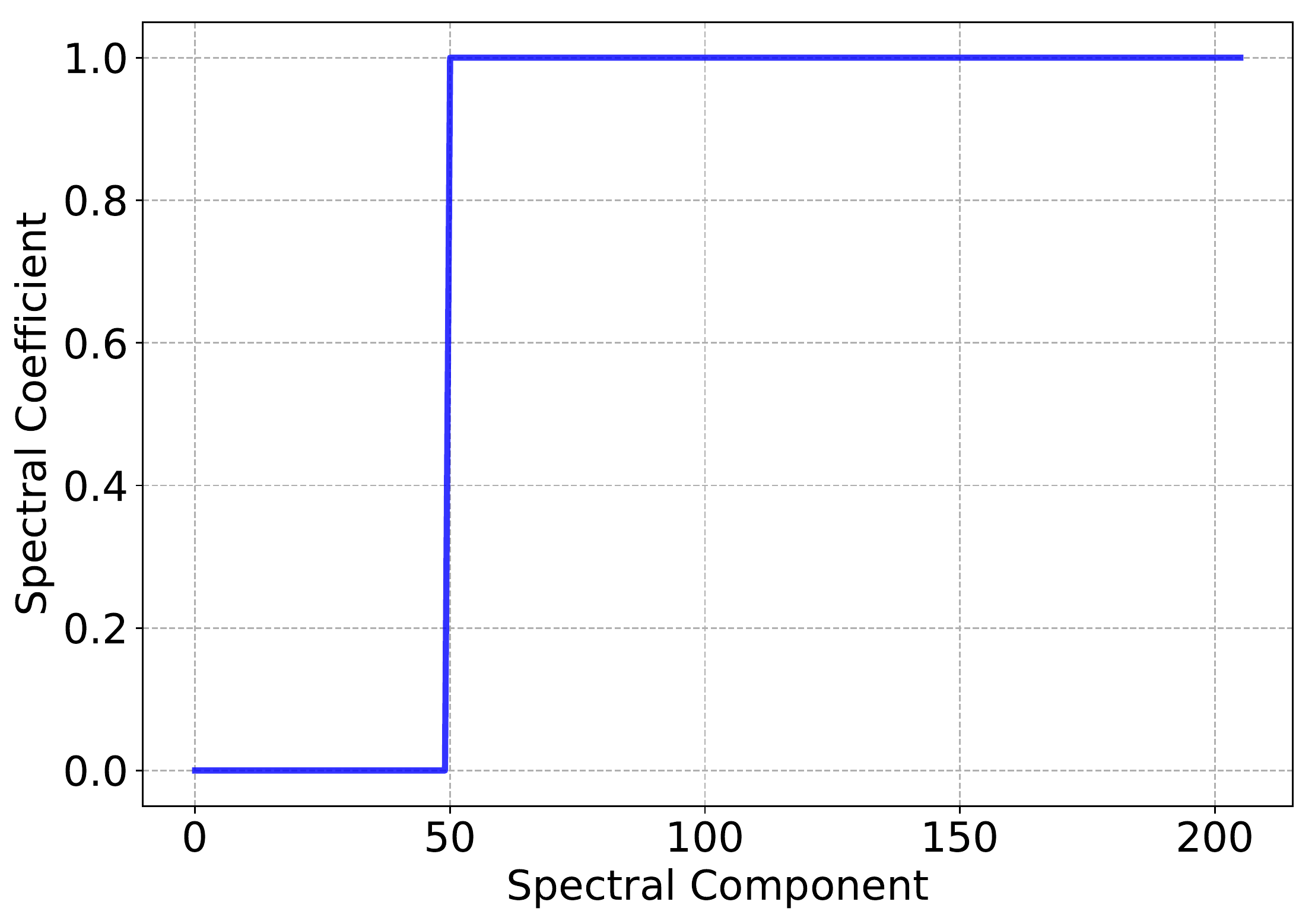}}
	\hfill
	\subfigure[Reconstructed graph signal]{\includegraphics[width=0.24\linewidth]{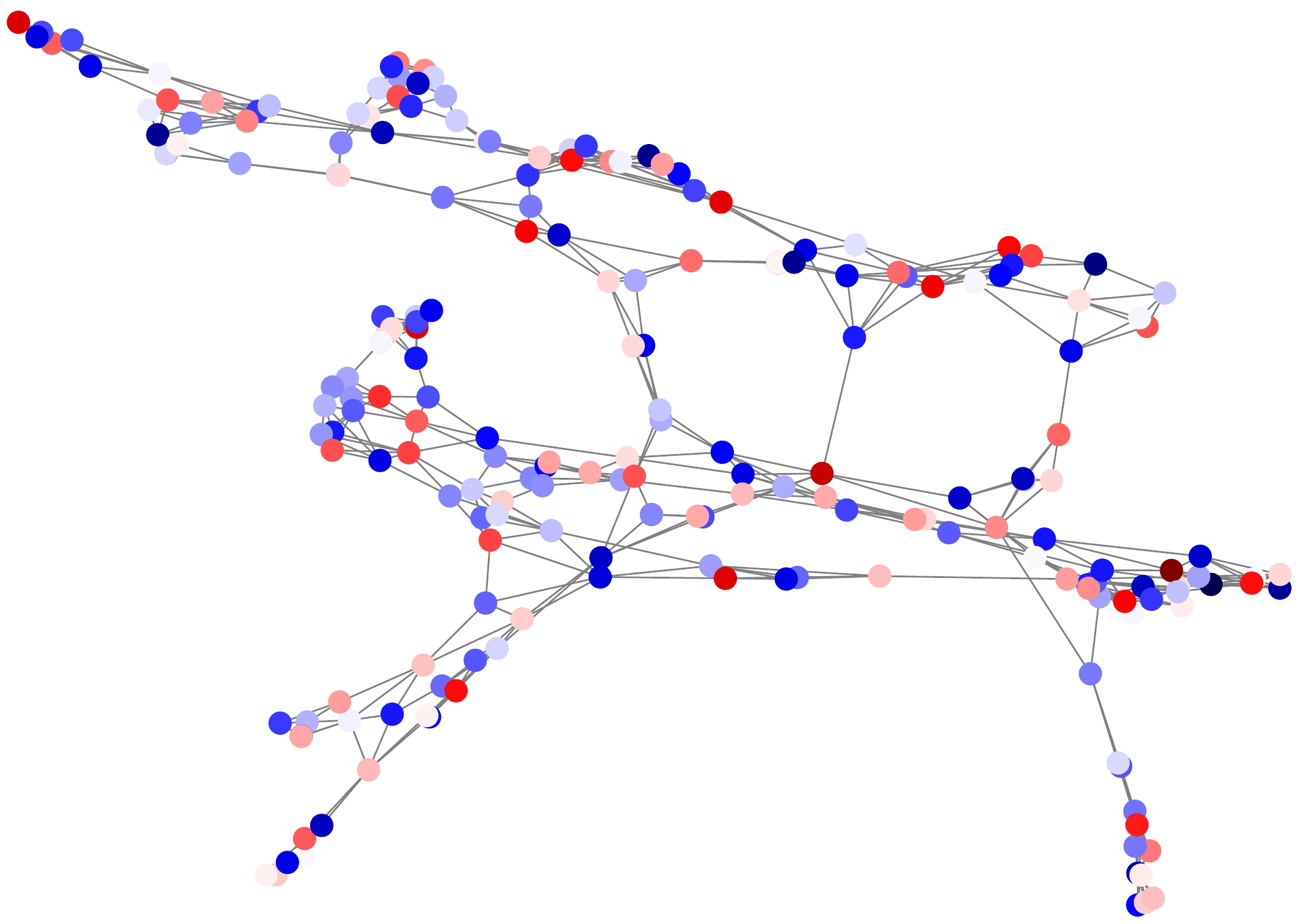}}
	\hfill
	\subfigure[High Reliability]{\includegraphics[width=0.24\linewidth]{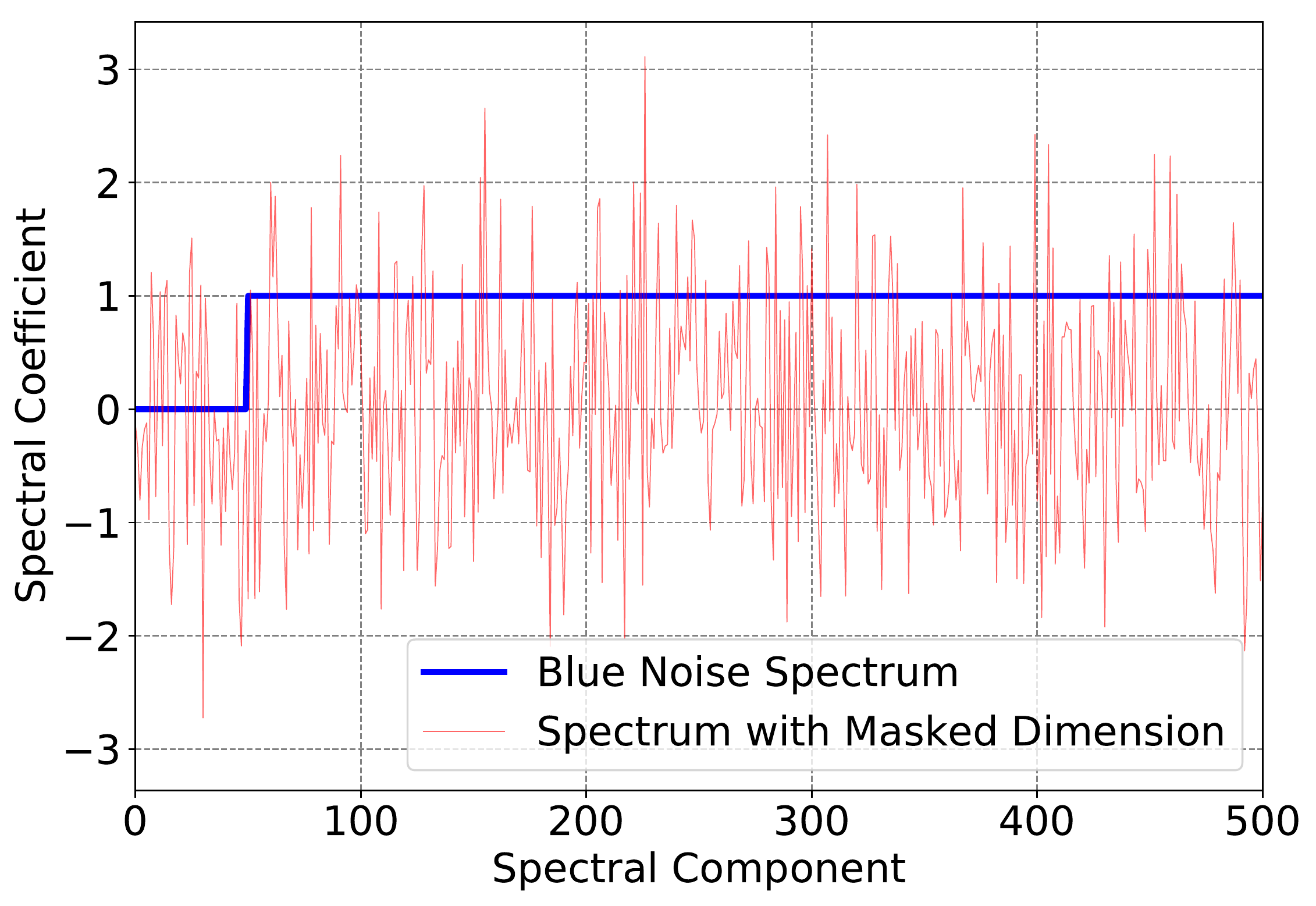}}
	\subfigure[Low Reliability]{\includegraphics[width=0.24\linewidth]{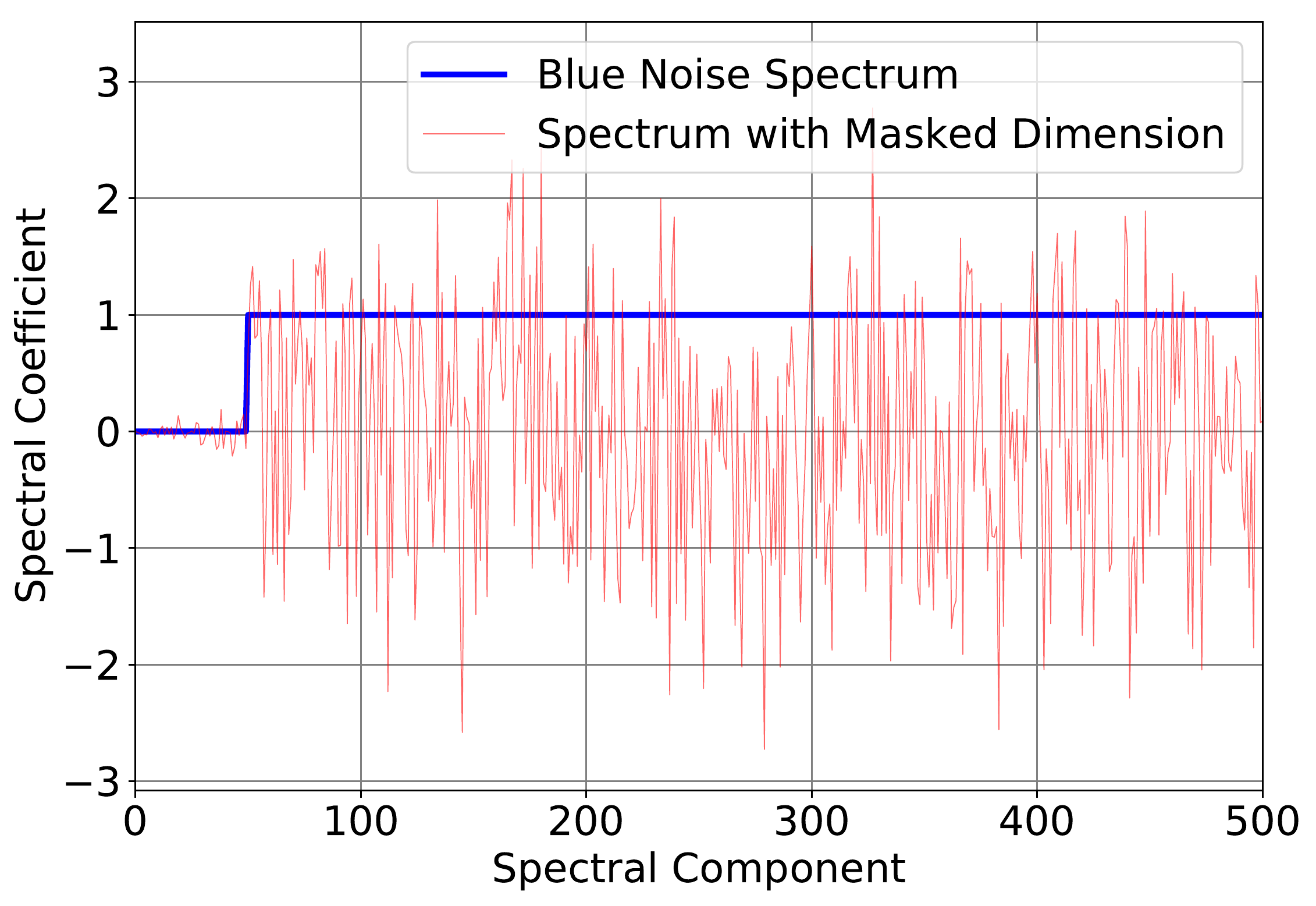}}
	\caption{(a)-(b) Designing a graph signal with blue noise spectrum., (c)-(d) Proposed importance measure for feature selection.}
	\label{fig:spec}
	\vspace{-0.1in}
\end{figure*}

Given a graph $\mathcal{G}$, we can define a graph signal $\mathbf{s}$, a numerical function indexed by the nodes $\mathcal{V}$, as follows: $\mathbf{s} = [s_1,s_2 \cdots s_N]^T; \forall s_i \in \mathbb{R}.$
For example, an image can be represented as pixels defined on a $2-$D regular lattice graph, and in this case, the pixel values form the graph signal. Following~\cite{sandryhaila2013discrete}, we define the \textit{graph shift} operator, akin to the \emph{time-shift} operator in classical signal processing. With the graph shift operation, the signal $s_i$ indexed by the node $v_i$ can be transformed as a weighted linear combination of the signal values at the neighboring nodes: 
\begin{equation}
\tilde{s}_i = \sum_{j = 1}^N \mathbf{A}_{i,j}s_j \implies \tilde{\mathbf{s}} = \mathbf{A} \mathbf{s}.
\label{eqn:grshift}
\end{equation}Here, the adjacency matrix is directly used to define the graph shift. Alternative choices include the transition matrix $\mathbf{D}^{-1}\mathbf{A}$, or the normalized graph Laplacian $\mathbf{L}$, which is used in this paper. 

\noindent \textbf{Graph Fourier Transform}: Performing spectral decomposition of a signal space $\mathcal{S}$ is at the core of the proposed approach. In general, spectral decomposition of a signal space corresponds to identifying subspaces that are invariant to the choice of filtering, i.e. the filtered version of a signal from subspace $\mathcal{S}_k$ still lies in that subspace. The set of generalized eigenvectors of the graph Laplacian, $\mathbf{L} = \mathbf{U} \mathbf{\Lambda} \mathbf{U}^T$, where $\mathbf{U} \in \mathbb{R}^{N\times N}$, is referred to as the graph Fourier basis. Consequently, decomposition of a signal $\mathbf{s} \in \mathcal{S}$ corresponds to computing its expansion in the graph Fourier basis: $\mathbf{s} = \mathbf{U} \hat{\mathbf{s}}$, where the expansion coefficients can be computed as $\hat{\mathbf{s}} = \mathbf{U}^{-1} \mathbf{s}$.
This process is known as the Graph Fourier Transform (GFT), and the collection of coefficients $\hat{\mathbf{s}}$ is referred to as the \textit{spectrum}~\cite{chen2015discrete}. The ordered eigenvalues loosely represent frequencies of signal variation, with $\lambda_1$ to $\lambda_N$ representing the smallest to largest frequencies. In other words, larger signal variations between closely connected neighbors correspond to high frequencies, while smooth variations correspond to low frequencies. In this context, the graph filtering using a graph shift operator corresponds to a simple \emph{low-pass} filter.


\section{Proposed Approach}
\subsection{Blue Noise Spectrum}
In supervised feature selection approaches, predictability and uncertainty are two commonly used heuristics for ranking features. While the former metric measures how well a feature supports the overall prediction, the latter measures how much the prediction is bound to change when a feature is perturbed. In the context of graph signal analysis, a predictable signal is characterized by the smoothness property with respect to the neighborhoods. In other words, we expect a graph spectrum to be dominated by low frequency content when a signal is predictable in the domain considered. Similarly, a response function that is highly uncorrelated with the predictor variables manifests as a spectrum with majority of its energy concentrated at higher frequencies. In unsupervised scenarios, we argue that by choosing an appropriate graph signal, one can effectively measure feature importances in a similar way.

The core idea of the proposed approach is that using a noise spectrum with controlled characteristics, we determine which dimensions maximally alter the spectral characteristics of the signal when they are perturbed. In particular, we consider the form of perturbation where a chosen dimension is masked (set to zero or a constant). For a given graph $\mathcal{G}$, we propose to utilize the \textit{blue} noise spectrum for studying feature importance.


\begin{figure*}[t]
	\centering
	\subfigure[\textit{Breast Cancer - Wisconsin}]{\includegraphics[width=0.27\linewidth]{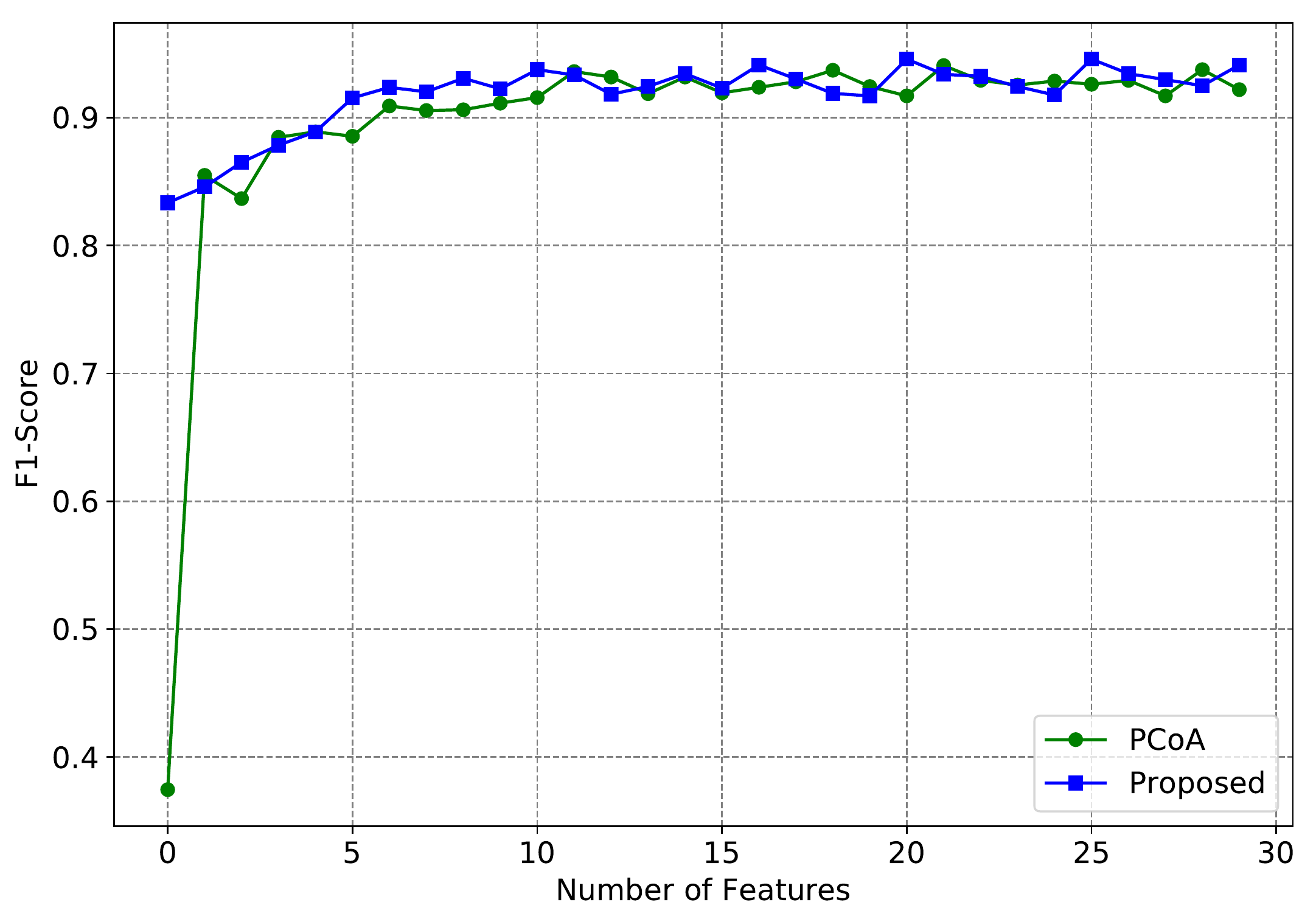}}
	\subfigure[\textit{Mfeat-Fourier}]{\includegraphics[width=0.27\linewidth]{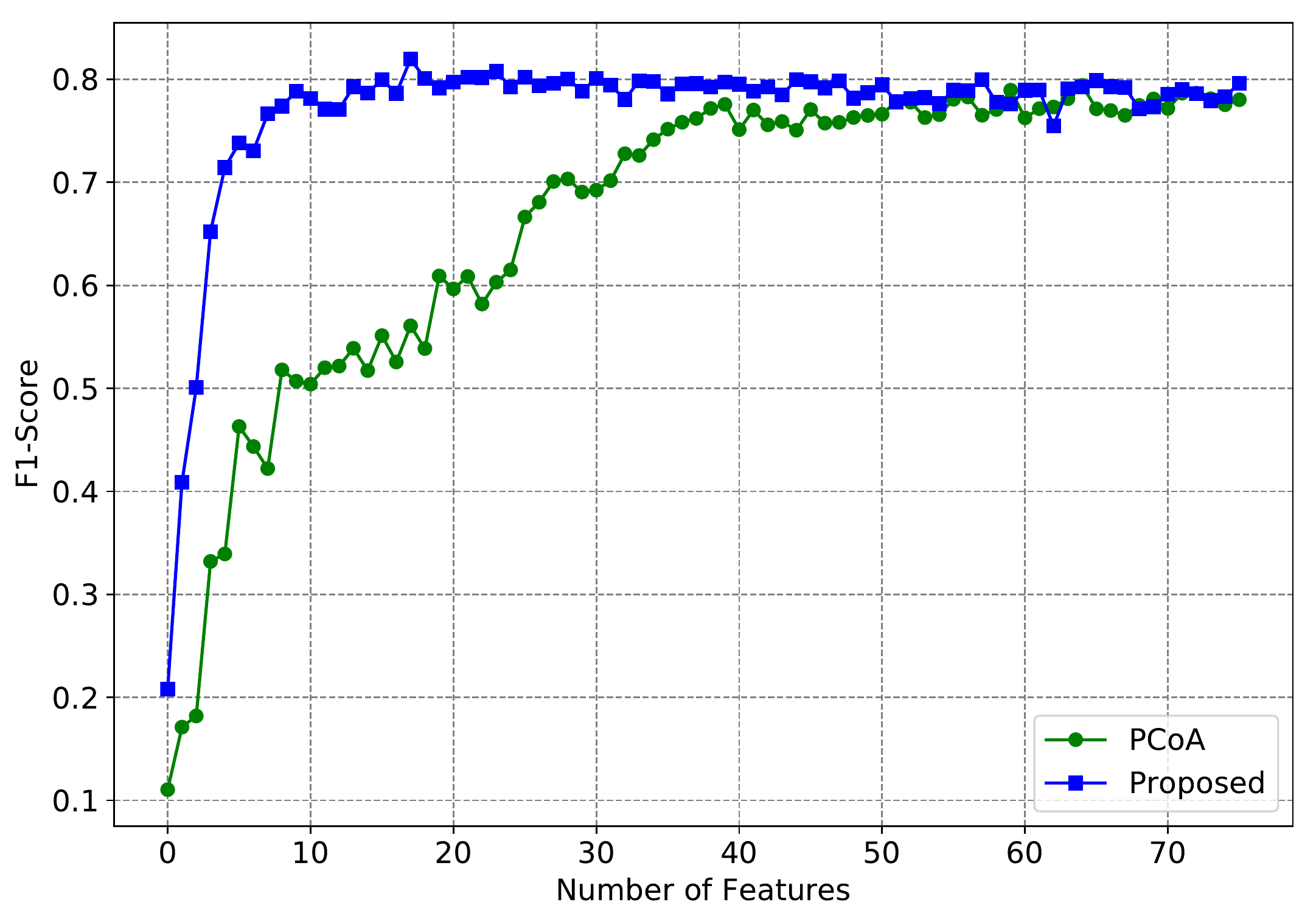}}
	\subfigure[\textit{Mfeat-pixel}]{\includegraphics[width=0.27\linewidth]{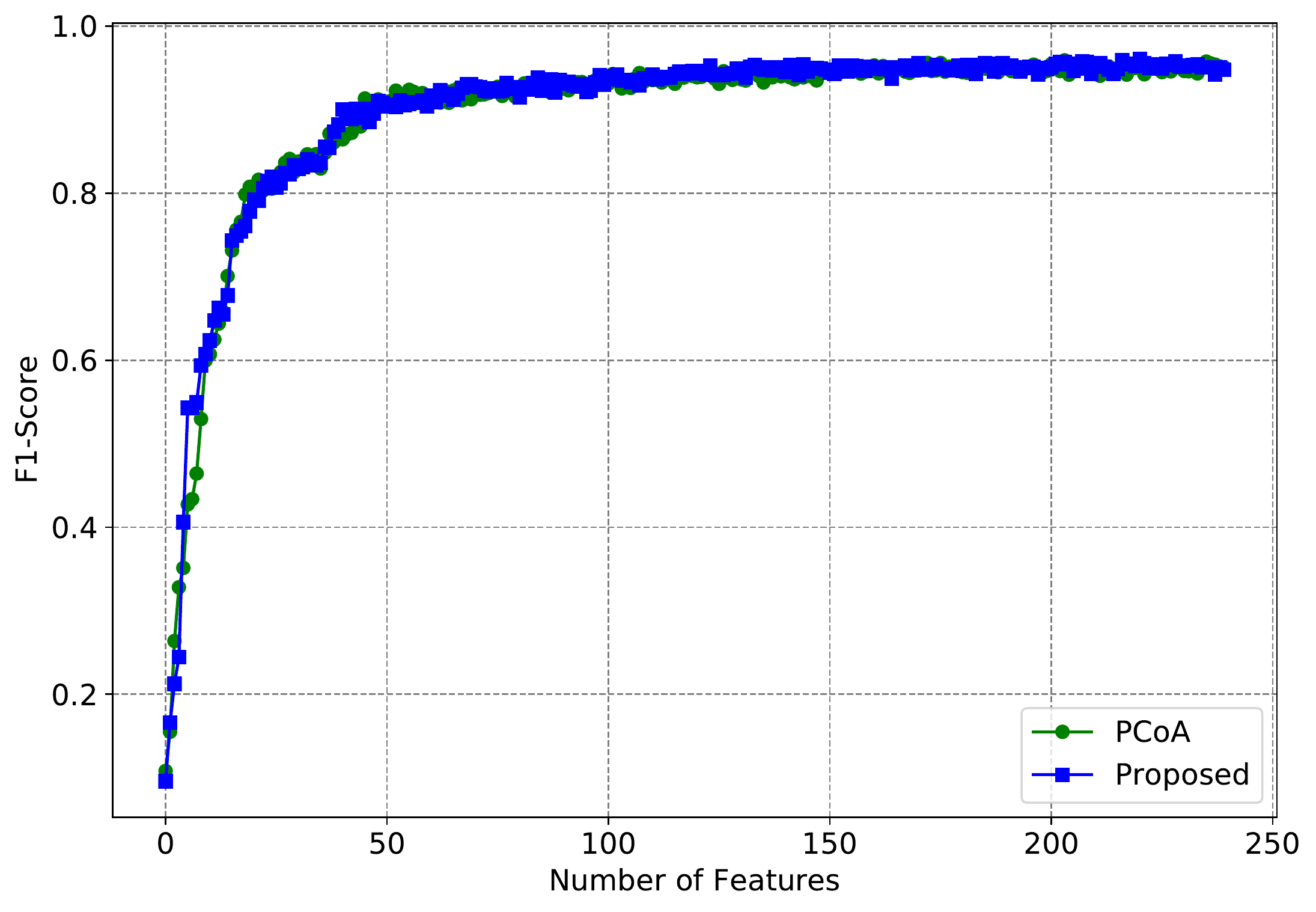}}
	\subfigure[\textit{Mfeat-factor}]{\includegraphics[width=0.27\linewidth]{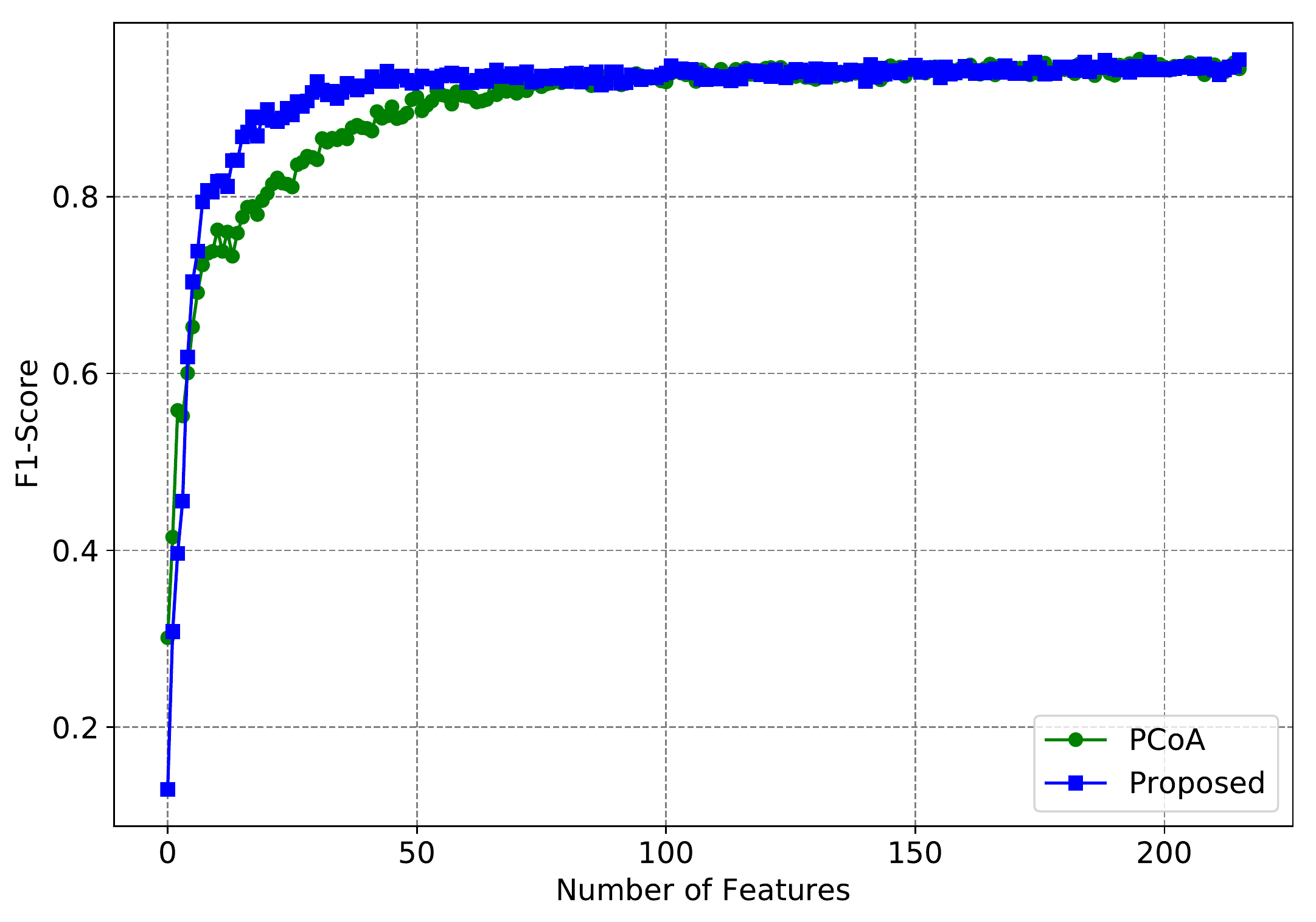}}
	\subfigure[\textit{Musk Clean1}]{\includegraphics[width=0.27\linewidth]{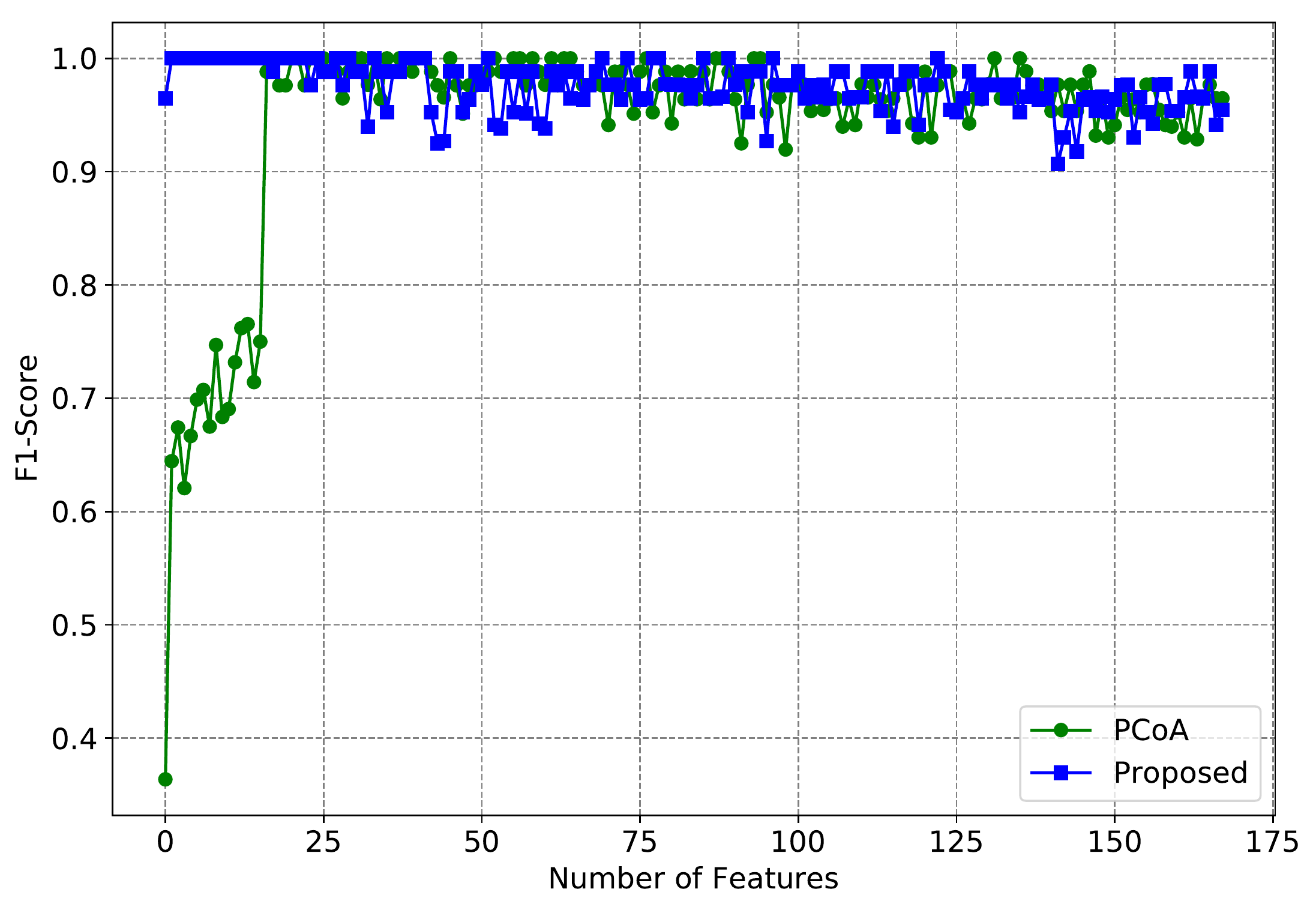}}
	\subfigure[\textit{Primate splice-junction gene sequences}]{\includegraphics[width=0.27\linewidth]{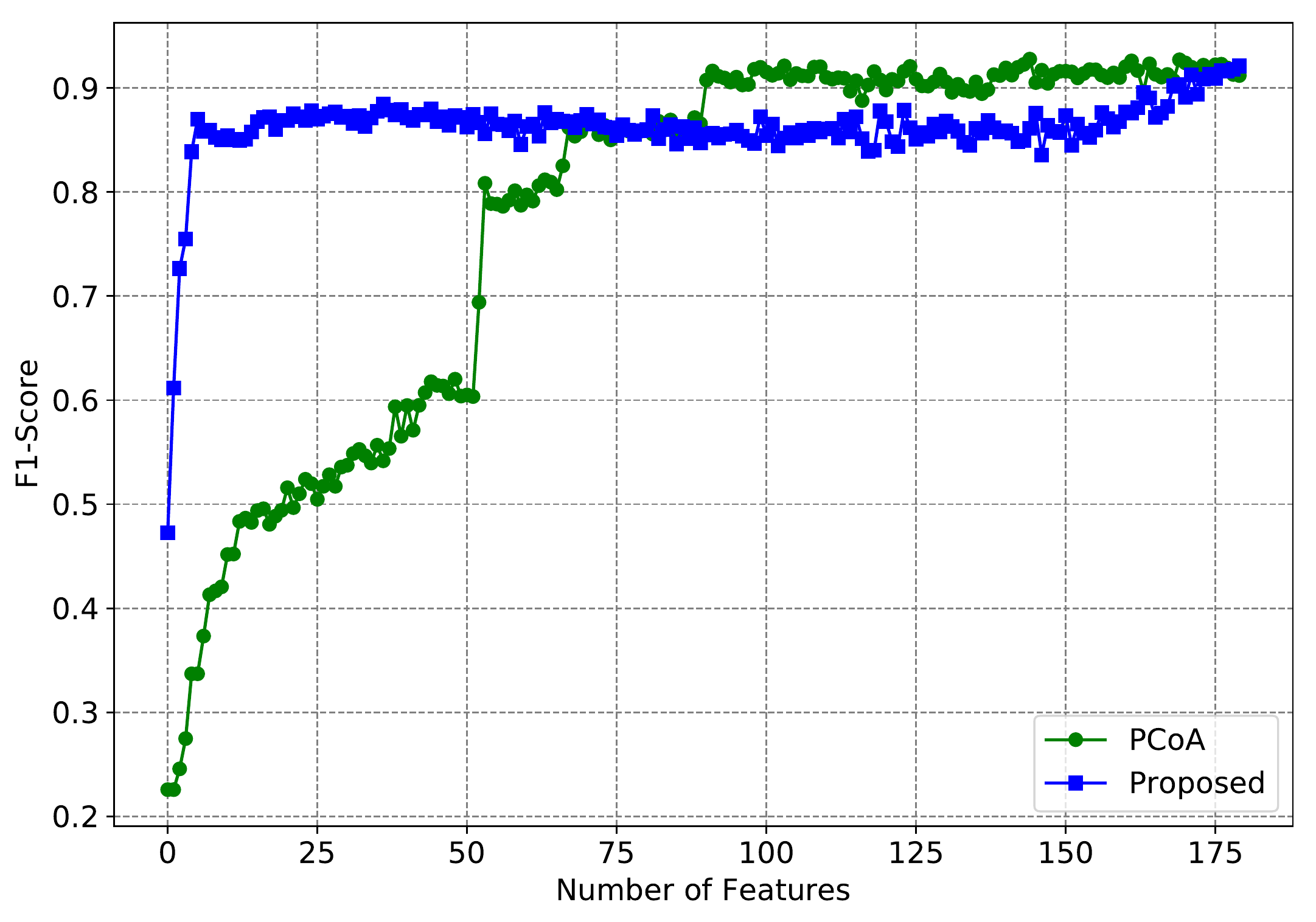}}
	\caption{Impact of feature selection on the classifier performance of datasets from the PMLB benchmark suite. We incrementally add features, ranked by the importance score, to the feature set and evaluate the validation performance of the resulting classifiers. We report the performance of \textit{Principal Coordinate Analysis} (PCoA) for comparison.}
	\label{fig:classify}
	\vspace{-0.1in}
\end{figure*}

Blue noise patterns have been regularly used in computer graphics for designing sampling distributions. In imaging problems, blue noise distributions~\cite{heck2013blue, kailkhura2016stair} are aimed at replacing visible aliasing artifacts with incoherent noise, and its properties are typically defined in the spectral domain. Formally, a blue noise power spectrum should satisfy the following two requirements: (a) the spectrum
should be close to zero in the low-frequency region, which indicates the frequencies that can be represented with no aliasing; (b) the spectrum should be a constant in mid and high-frequency regions to reduce the risk of aliasing. The low frequency band with minimal energy is referred to as the \textit{zero region}. Hence, we define a blue noise graph spectrum as:
\begin{equation}
\hat{s}_k^{b} = \left\{ \begin{array}{rll}
0  & \mbox{if}\ k\leq k_0, \\
1  & \mbox{if}\ k>k_0.
\end{array}\right. 
\label{steppsd}
\end{equation}Here, $k\leq k_0$ denotes the range of low-frequency spectral components that will have zero energy. In our context, we expect a signal with a blue noise graph spectrum to have no smoothness with respect to $\mathcal{G}$ and have an equally likely chance of observing all frequencies larger than $k_0$. Using the inverse GFT, we can reconstruct the signal corresponding to the blue noise spectrum as $\mathbf{s}^{b} = \mathbf{U} \hat{\mathbf{s}}^{b}$. Figures \ref{fig:spec}(a) and \ref{fig:spec}(b) illustrate the blue noise spectrum (with $k_0=50$) and its corresponding inverse Fourier transform for an example case.

\subsection{Measuring Feature Importances}
In order to measure the importance of different dimensions in a high-dimensional dataset $\mathbf{X}$ with $N$ samples and the feature set $\mathcal{D}$, we first construct a $k-$nearest neighbor graph $\mathcal{G}$ and build a signal with blue noise spectrum. Subsequently, we mask one dimension $d \in \mathcal{D}$ at a time and rebuild the neighborhood graph $\mathcal{G}_d$ with the modified data. By computing the GFT using the new set of basis functions, $\mathbf{U}_d$, we obtain the modified spectrum for the blue noise signal. A feature dimension $d$ is considered to be more important when masking that dimension introduces significant low-frequency components into the spectrum. In other words, an otherwise non-smooth signal becomes more smooth when the domain is altered by masking one of the relevant feature dimensions. For example, Figure \ref{fig:spec}(c) shows the modified spectrum (along with the original blue noise spectrum) when a relevant feature is masked. Whereas, as seen in Figure \ref{fig:spec}(d), perturbing an irrelevant feature still results in nearly zero low-frequency content. In contrast to existing masking techniques~\cite{dadkhahi2016masking}, this approach does not sequentially augment the set of selected features, but instead studies each dimension one at a time with respect to the blue noise spectrum. Consequently, this can be entirely parallelized and made scalable to even higher dimensional datasets. 

Finally, we define an importance score that is used to rank the dimensions in $\mathcal{D}$. Denoting the original blue noise spectrum and the spectrum with a masked dimension as $\hat{\mathbf{s}}^b$ and $\hat{\mathbf{s}}^b_d$ respectively, we define the importance score as follows:
\begin{equation}
\gamma_{d} = \left\|\hat{\mathbf{s}}^b[1:k_0]\right\|_2^2 - \left\|\hat{\mathbf{s}}^b_d[1:k_0] \right\|_2^2.
\label{eqn:score}
\end{equation}Here, we measure the difference in total energies at low-frequencies, in order to quantify the amount of smoothness in the modified spectrum. The parameter $k_0$ corresponds to the zero region in the definition of the blue noise spectrum. Interestingly, from our experiments, we find that the resulting feature ranking is not very sensitive to the choice of $k_0$, and hence, we fixed $k_0 = 100$ in all our studies.


\section{Experiments}
\subsection{Impact of Feature Selection on Classifier Design}
In this experiment, we evaluate the proposed approach by using the selected subset to design classifiers, which is especially important in small data scenarios. Though including a large number of features can provide flexibility, model robustness can suffer when working with datasets that have a lot of dimensions, but very few samples. Consequently, unsupervised feature selection can be an effective pre-processing step prior to model design. For all datasets considered in this experiment, we used the extremely randomized trees ($20$ estimators) model, and we report results from 5-fold cross-validation.

\noindent \textbf{Data}: For this experiment, we considered $6$ datasets from the Penn Machine Learning Benchmark (PMLB), which encompasses a wide range
of existing benchmark datasets for ML algorithms~\cite{olson2017pmlb}: (i) \textit{Breast Cancer - Wisconsin} for diagnosis of breast tissues with 569 samples in 30 dimensions; Three datasets consisting of features from 2000 handwritten digits extracted from a collection of Dutch utility maps - (ii) \textit{Mfeat-Fourier} consisting 76 Fourier coefficients of the character shapes; (iii) \textit{Mfeat-pixel} comprised of 240 pixel average values for $2 \times 3$ windows; (iv) \textit{Mfeat-factor} consisting 216 feature correlation values; (v) \textit{Musk Clean1} dataset for predicting whether a molecule is musk or non-musk with 476 instances in 168 dimensions; and (vi) \textit{Primate splice-junction gene sequences} data consisting 3186 data points (splice junctions) described by 180 binary indicator variables.

\begin{figure}[t]
	\centering
	\includegraphics[width=0.8\linewidth]{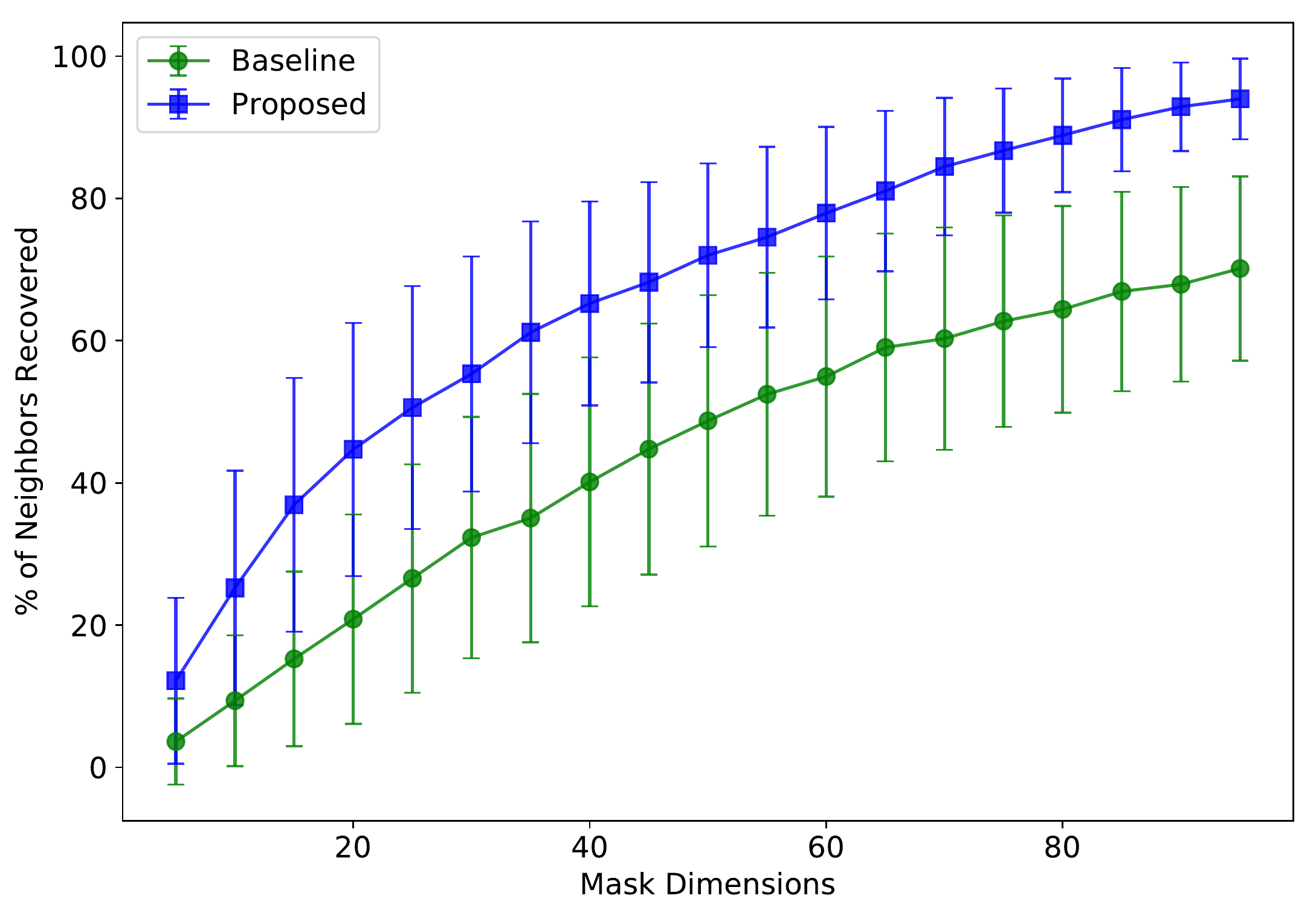}
	
	\caption{Image Masking Results - Performance of the masks inferred using the proposed approach in preserving the local topology of high-dimensional data. For comparison, we show results obtained using the dimension masking approach in \cite{dadkhahi2016masking}.}
	\label{fig:neigh}
	\vspace{-0.1in}
\end{figure}

\noindent \textbf{Baseline}: For comparison, we used principal coordinate analysis (PCoA), an adaptation of principal component analysis for unsupervised dimension selection~\cite{dadkhahi2015image}. The core idea of PCoA is to identify the canonical basis vectors, in lieu of orthogonal linear projections in PCA, that span the canonical subspace which maximally describes the variations of the data:
$$\omega_i = \max_{i \in \mathcal{D}} \sum_{n=1}^N \left(\mathbf{x}_n[i] - \bar{\mathbf{x}}[i]\right)^2,$$where $\mathcal{D}$ is the set of input features and $\bar{\mathbf{x}}$ denotes the mean of the dataset. Using the above strategy, we obtain the feature set $\Omega = \{\omega_i\}$ by greedily selecting features by ranking them based on their variances across the dataset.

\noindent \textbf{Results}: Figure \ref{fig:classify} shows the classification performance (measured using the F1-score) obtained using the selected features, by incrementally adding one feature at a time. For the proposed approach, we ranked the features based on the importance scores defined in (\ref{eqn:score}). We can see that the proposed approach consistently outperforms the baseline technique, particularly with a small number of features. For example, with the \textit{Mfeat-Fourier} dataset, the proposed approach reaches an F1-score of 0.8 with just 10 features, while PCoA requires 40 dimensions to achieve the same validation performance. Interestingly, with these HD datasets, adding more features does not always lead to improved performances. For example, the \textit{Musk Clean1} dataset requires only $3$ features from our approach to reach an F1-score of 0.99, however the performance obtained using all $168$ features is lower. This clearly evidences the challenge of dealing with HD feature spaces in small datasets.

\subsection{Application: Image Masking}
An important application of feature selection is in designing masking patterns for building image sensors. The goal is to acquire only a subset of pixel locations for a given distribution of images, such that it provides enough information to recover the local topology, i.e. neighboring images. Hence, the metric we use for evaluation is the percentage of neighbors recovered using only the subset of selected features, which are individual pixels in this case. For all cases, the number of neighbors in the full-dimensional space was fixed at $20$.

\begin{figure}[t]
	\centering
	\includegraphics[width=0.85\linewidth]{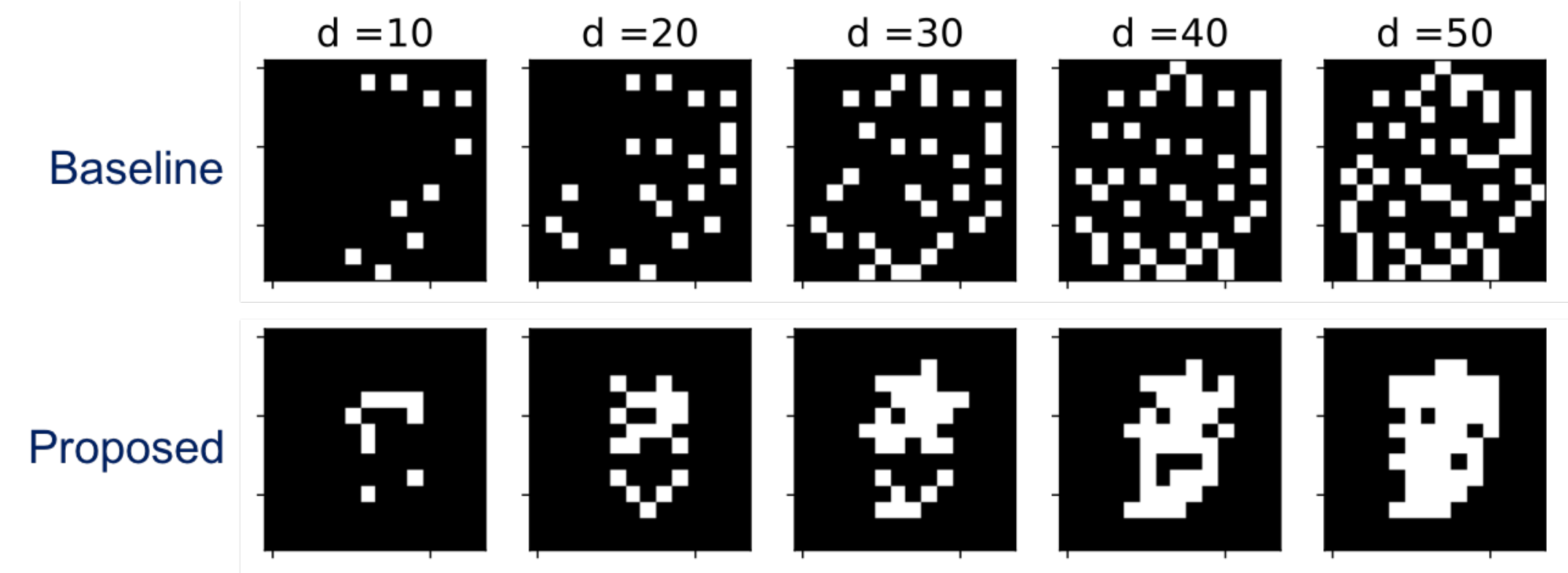}
	
	\caption{Image Masking Results - Masks learned using the proposed approach and the baseline technique from \cite{dadkhahi2016masking}. d corresponds to the number of dimensions.}
	\label{fig:mask}
	\vspace{-0.1in}
\end{figure}

\noindent \textbf{Data}: For this experiment, we used a subset of the MNIST digits dataset, by considering $1000$ randomly selected images, with $100$ images for each digit, in order to infer the masking pattern, and used an unseen set of $5000$ images for evaluation. 

\noindent \textbf{Baseline}: The dimension masking algorithm proposed in \cite{dadkhahi2016masking} provides a state-of-the-art approach for solving the problem of identifying critical dimensions that preserve the neighborhood structure. In particular, we use the generalization of Isomap for dimension selection, and evaluate its performance in comparison to the proposed approach.

\noindent \textbf{Results}: From Figure \ref{fig:neigh}, we can see that the proposed approach achieves significant improvements ($~20\%$) in the recovery performance as compared to the state-of-the-art image masking technique. Furthermore, the masking patterns for the two approaches are illustrated in Figure \ref{fig:mask}, which clearly evidence the effectiveness of the graph analysis approach in localizing the regions where maximal information can be acquired.

\section{Conclusion}
In this paper, we presented an unsupervised approach to dimension selection based on graph signal analysis of a blue noise spectrum. With the help of applications in supervised learning and image masking, we showed its efficacy over existing techniques in terms of its characterization of feature spaces, supervisory performance, and its ability to be parallelized and scaled. In the future, it would be interesting to (a) evaluate the approach on datasets with many more dimensions, (b) extend it to applications such as uncertainty quantification, and label propagation, and (c) study its robustness with respect to changes in the formulation of the nearest neighbors graphs.

\bibliographystyle{IEEEbib}
\bibliography{refs}

\end{document}